\documentclass{article}
\usepackage{amsmath,graphicx,spconf}
\usepackage{amsthm}
\usepackage[T1]{fontenc}
\usepackage{hyperref, times, cite}
\usepackage{url, datetime}
\usepackage{enumitem,kantlipsum}
\usepackage{mathtools}
\usepackage{bm}
\usepackage{cool}
\usepackage{amssymb,amsfo nts,amsmath,amsthm,amscd,dsfont,mathrsfs}
\usepackage{graphicx,float,psfrag,epsfig,amssymb}
\usepackage{wrapfig}
\usepackage{relsize}
\usepackage{color}
\usepackage{pict2e}
\usepackage{algorithm}
\usepackage{algorithmic}
\usepackage{caption}
\usepackage{nameref}
\usepackage{enumerate}
\usepackage{stackrel}

\usepackage{caption}
\usepackage{subcaption}
\title{Sparse-Group Log-Sum Penalized Graphical Model Learning For Time Series}
\name{{\em Jitendra K.\ Tugnait}  \thanks{This work was supported by NSF Grant ECCS-2040536. Author's email: tugnajk@auburn.edu} \vspace*{-0.1in}}
\address{Department of Electrical \& Computer Engineering\\
Auburn University, Auburn, AL 36849, USA}
\begin{document}
\ninept

\maketitle

\setcounter{footnote}{-1}
\def\thefootnote{\fnsymbol{footnote}}
\renewcommand{\algorithmicrequire}{\textbf{Input:}}
\renewcommand{\algorithmicensure}{\textbf{Output:}}

\begin{abstract}
We consider the problem of inferring the conditional independence graph (CIG) of a high-dimensional stationary multivariate Gaussian time series. A sparse-group lasso based frequency-domain formulation of the problem has been considered in the literature where the objective is to estimate the sparse inverse power spectral density (PSD) of the data. The CIG is then inferred from the estimated inverse PSD. In this paper we investigate use of a sparse-group log-sum penalty (LSP) instead of sparse-group lasso penalty. An alternating direction method of multipliers (ADMM) approach for iterative optimization of the non-convex problem is presented. We provide sufficient conditions for local convergence in the Frobenius norm of the inverse PSD estimators to the true value. This results also yields a rate of convergence. We illustrate our approach using numerical examples utilizing both synthetic and real data.
\end{abstract}


\section{Introduction} \label{intro}

Graphical models are an important and useful tool for analyzing multivariate data \cite{Lauritzen1996}. A central concept is that of conditional independence. Given a collection of random variables, one wishes to assess the relationship between two variables, conditioned on the remaining variables. 
Consider a graph ${\cal G} = \left( V, {\cal E} \right)$ with a set of $p$ vertices (nodes) $V = \{1,2, \cdots , p\} =[p]$, and a corresponding set of (undirected) edges ${\cal E} \subseteq [p] \times [p]$. Also consider a stationary (real-valued), zero-mean,  $p-$dimensional multivariate Gaussian time series ${\bm x}(t)$, $t=0, \pm 1, \pm 2, \cdots $, with $i$th component $x_i(t)$, and correlation (covariance) matrix function ${\bm R}_{xx}( \tau ) = \mathbb{E} \{ {\bm x}(t + \tau) {\bm x}^T(t ) \}$, $ \tau = 0, \pm 1,  \cdots  $. Given $\{ {\bm x}(t) \}$, in the corresponding graph ${\cal G}$, each component series $\{ x_i(t) \}$ is represented by a node ($i$ in $V$), and associations between components $\{ x_i (t) \}$ and $\{ x_j(t) \}$ are represented by edges between nodes $i$ and $j$ of ${\cal G}$. In a conditional independence graph (CIG), there is no edge between nodes $i$ and $j$ if and only if (iff) $x_i(t)$ and $x_j(t)$ are conditionally independent given the remaining $p$-$2$ scalar series $x_\ell(t)$, $\ell \in [p]$, $\ell \neq i$, $\ell \neq j$. 

A key insight in \cite{Dahlhaus2000} was to transform the series to the frequency domain and express the graph relationships in the frequency domain. Denote the power spectral density (PSD) matrix of $\{ {\bm x}(t) \}$ by ${\bm S}_x(f)$, where ${\bm S}_x(f) = \sum_{\tau = -\infty}^{\infty}  {\bm R}_{xx}( \tau ) e^{-j 2 \pi f \tau}$. In \cite{Dahlhaus2000} it was shown that conditional independence of two time series components given all other components of the time series, is encoded by zeros in the inverse PSD, that is, $\{ i,j \} \not\in {\cal E}$ iff the $(i,j)$-th element of ${\bm S}_x(f)$,  $[{\bm S}_x^{-1}(f)]_{ij} = 0$ for every $f$. Hence one can use estimated inverse PSD of observed time series to infer the associated graph.

Graphical models were originally developed for random vectors \cite[p.\ 234]{Eichler2012}. Such models have been extensively studied, and found to be useful in a wide variety of applications \cite{Danaher2014, Friedman2004, Lauritzen2003, Meinshausen2006, Mohan2014}. Graphical modeling of real-valued time-dependent data (stationary time series) originated with \cite{Brillinger1996}, followed by \cite{Dahlhaus2000}. Nonparametric approaches for graphical modeling of real time series in high-dimensional settings ($p$ is large and/or sample size $n$ is of the order of $p$) have been formulated in frequency-domain in \cite{Jung2014, Jung2015b} using a neighborhood regression scheme, and in the form of group-lasso penalized log-likelihood in frequency-domain in \cite{Jung2015a}. Sparse-group lasso penalized log-likelihood approach in frequency-domain has been considered in \cite{Tugnait18c, Tugnait20}. 

In this paper we consider a sparse-group log-sum penalty (SGLSP) instead of sparse-group lasso (SGL) penalty (\cite{Tugnait18c, Tugnait20}) to regularize the problem, motivated by \cite{Candes2008}. For sparse solutions, ideal penalty is $\ell_0$ which is non-convex and the problem is usually impossible to solve. So one relaxes the problem using $\ell_1$ (lasso) penalty which is convex. \cite{Candes2008} notes that a key difference between the $\ell_1$ and $\ell_0$ norms is the dependence on magnitude: ``larger coefficients are penalized more heavily in the $\ell_1$ norm than smaller coefficients, unlike the more democratic penalization of the $\ell_0$ norm.'' Their solution to rectify this imbalance, is iterative reweighted $\ell_1$ minimization, and to construct an analytical framework, \cite{Candes2008} suggests the log-sum penalty.
We present an ADMM approach for iterative optimization of the non-convex problem. We provide sufficient conditions for consistency of a local estimator of inverse PSD. We illustrate our approach using numerical examples utilizing both synthetic and real data. Synthetic data example shows that our SGLSP approach significantly outperforms the SGL and other approaches in correctly detecting the graph edges.

{\em Notation}: $|{\bm A}|$ and ${\rm tr}({\bm A})$ denote the determinant and the trace
of the square matrix ${\bm A}$, respectively. $[{\bm B}]_{ij}$ denotes the $(i,j)$-th element of ${\bm B}$, and so does $B_{ij}$. ${\bm I}$ is the identity matrix.
The superscripts $\ast$ and $H$ denote the complex conjugate and the Hermitian (conjugate transpose) operations, respectively. The notation ${\bm x} \sim {\mathcal N}_c( {\bm m}, \bm{\Sigma})$ denotes a random vector  ${\bm x}$ that is circularly symmetric (proper) complex Gaussian with mean ${\bm m}$  and covariance $\bm{\Sigma}$.
\vspace*{-0.1in}
\section{Sparse-Group Lasso Penalized Negative Log-Likelihood} 
Given ${\bm x}(t)$ for $t=0, 1,2, \cdots , n-1$. Define the (normalized) DFT ${\bm d}_x(f_m)$ of ${\bm x}(t)$, ($j = \sqrt{-1}$, $f_m = m/n$), over $m=0,1, \cdots , n-1$ as ${\bm d}_x(f_m) =  \frac{1}{\sqrt{n}} \sum_{t=0}^{n-1} {\bm x}(t) \exp \left( - j 2 \pi f_m t \right)$.
It is established in \cite{Tugnait19d} that the set of random vectors $\{{\bm d}_x(f_m)\}_{m=0}^{n/2}$ is a sufficient statistic for any inference problem based on dataset $\{ {\bm x}(t) \}_{t=0}^{n-1}$. Suppose  ${\bm S}_x(f_k)$ is locally smooth, so that ${\bm S}_x(f_k)$ is (approximately) constant over $K=2m_t+1$ consecutive frequency points $f_m$s. Pick $M =  \left\lfloor (\frac{n}{2}-m_t-1)/K \right\rfloor$ and
\begin{align} 
  \tilde{f}_k = & \frac{(k-1)K+m_t+1}{n}, \;\; \quad k=1,2, \cdots , M, \label{window} 
\end{align}
yielding $M$ equally spaced frequencies $\tilde{f}_k$ in the interval $(0,0.5)$. By local smoothness
\begin{align}  
   {\bm S}_x(\tilde{f}_{k,\ell}) = & {\bm S}_x(\tilde{f}_k) \,  
	\; \mbox{for  } \ell = -m_t, -m_t+1, \cdots , m_t, \label{eqth1_160}  \\
	\mbox{ where } \;  &
	  \tilde{f}_{k,\ell} = \frac{(k-1)K+m_t+1 + \ell}{n}.
\end{align}
It is known (\cite[Theorem 4.4.1]{Brillinger}) that asymptotically (as $n \rightarrow \infty$), ${\bm d}_x(f_m)$, $m=1,2, \cdots , (n/2)-1$, ($n$ even), are independent proper, complex Gaussian ${\mathcal N}_c( {\bf 0}, {\bm S}_x(f_m))$ random vectors, respectively, provided all elements of ${\bm R}_{xx}( \tau )$ are absolutely summable. Denote the joint probability density function of ${\bm d}_x(f_m)$, $m=1,2, \cdots , (n/2)-1$, as $f_{{\bm D}}({\bm D})$. 

We wish to estimate inverse PSD matrix $\bm{\Phi}_k := {\bm S}_x^{-1}(\tilde{f}_k)$. In terms of $\bm{\Phi}_k$, we have the log-likelihood \cite{Tugnait18c} (up to some constant)
\begin{align} 
    \ln & f_{{\bm D}}({\bm D})  \propto - G(\{ \bm{\Phi} \}, \{ \bm{\Phi}^\ast \}) \\
					&	:= \sum_{k=1}^M \frac{1}{2} \left[( \ln |\bm{\Phi}_k| +\ln |\bm{\Phi}_k^\ast| )
		        - {\rm tr} \left( \hat{\bm S}_k \bm{\Phi}_k  +
						     \hat{\bm S}_k^\ast \bm{\Phi}_k^\ast \right)  \right]    	 \label{eqth2_10}
\end{align}
where the PSD estimator using unweighted frequency-domain smoothing is $\hat{\bm S}_k = (1/K) \sum_{\ell= - m_t}^{m_t} {\bm d}_x(\tilde{f}_{k,\ell}) {\bm d}_x^H(\tilde{f}_{k,\ell})$.
In the high-dimension case of $K < p(p-1)/2$ (\# of unknowns in ${\bm S}_x^{-1}(\tilde{f}_k) )$), one may need to use penalty terms to enforce sparsity and to make the problem well-conditioned. Imposing a sparse-group sparsity constraint (cf.\ \cite{Friedman2010a, Friedman2010b, Danaher2014}), \cite{Tugnait18c} minimizes a penalized version of negative log-likelihood w.r.t.\ $\{ \bm{\Phi} \}$
\begin{align}  
   &L_{SGL} (\{ \bm{\Phi} \})  =  G(\{ \bm{\Phi} \}, \{ \bm{\Phi}^\ast \}) + \bar{P}(\{ \bm{\Phi} \}) , \label{eqth2_20} \\
	 &  \bar{P} (\{ \bm{\Phi} \} )   =  \bar{\lambda}_1 \, \sum_{k=1}^M \; \sum_{i \ne j}^p 
		  \Big| [ {\bm{\Phi}}_k ]_{ij} \Big|  
		    + \bar{\lambda}_2 \, \sum_{ i \ne j}^p \; \| {\bm{\Phi}}^{(ij)} \|  \label{eqth2_20b} \\
 &\mbox{where } \; {\bm{\Phi}}^{(ij)} := [ [{\bm{\Phi}}_1 ]_{ij} \; [{\bm{\Phi}}_2 ]_{ij} \; \cdots \; [{\bm{\Phi}}_M ]_{ij}]^\top
     \in \mathbb{C}^M  \label{eqth2_20c}
\end{align}
and $\bar{\lambda}_1, \, \bar{\lambda}_2 \ge 0$ are tuning parameters. An analysis of the properties of the minimizer $\{ \hat{\bm{\Phi}} \}$ of $L_{SGL}(\{ \bm{\Phi} \})$ is given in \cite{Tugnait20}. In \cite{Tugnait20}, $\bar{\lambda}_1 = \alpha \lambda$ and $\bar{\lambda}_2 = (1-\alpha) \lambda$ with $\lambda > 0$, and $\alpha \in [0,1]$ providing a convex combination of lasso and group-lasso penalties \cite{Friedman2010a, Friedman2010b}.

\section{Proposed Sparse-Group Log-Sum Penalized Negative Log-Likelihood}
With $0 < \epsilon \ll 1$, following \cite{Candes2008}, define the log penalty for $\theta \in \mathbb{R}$,
\begin{align} 
   p_{\lambda}(\theta) = \lambda \ln \left(1+ |\theta|/\epsilon \right) \,  .
	\label{neq130a}
\end{align}
Replace $\bar{P} (\{ \bm{\Phi} \} )$ in (\ref{eqth2_20b}) with ${P} (\{ \bm{\Phi} \} )$, defined as
\begin{align}  
	 &  {P} (\{ \bm{\Phi} \} )   =  \sum_{k=1}^M \; \sum_{i \ne j}^p 
		  p_{\bar{\lambda}_1}( [{\bm{\Phi}}_k ]_{ij} )  
		    + \sum_{ i \ne j}^p \; p_{\bar{\lambda}_2}(\| {\bm{\Phi}}^{(ij)} \|) \, .  \label{eqth2_20d} 
\end{align}
Now replace (\ref{eqth2_20}) with (\ref{eqth2_20ee})
\begin{equation} 
   L_{LSP} (\{ \bm{\Phi} \})  =  G(\{ \bm{\Phi} \}, \{ \bm{\Phi}^\ast \}) 
	 + {P}(\{ \bm{\Phi} \}) , \label{eqth2_20ee}
\end{equation}
to define the sparse-group log-sum penalized log-likelihood function. Unlike $L_{SGL} (\{ \bm{\Phi} \})$, we now have a non-convex function of $\{ \bm{\Phi} \}$ in $L_{LSP} (\{ \bm{\Phi} \})$. 

As for the SCAD (smoothly clipped absolute deviation) penalty in \cite{Lam2009}, we solve the problem $\min_{ \bm{\Phi}_k \succ {\bm 0}, \; k=1, \cdots M } \, L_{LSP} (\{ \bm{\Phi} \})$ iteratively, where in each iteration, the problem is convex. Using $\partial p_{\lambda}(|\theta|) / \partial |\theta| = \lambda /(|\theta| + \epsilon)$, a local linear approximation to $p_{\lambda}(|\theta|)$ around $\theta_0$ yields
\begin{equation}
  p_{\lambda}(|\theta|) \approx P_{\lambda}(|\theta_0|) 
	 + \frac{\lambda}{|\theta_0| + \epsilon} (|\theta| - |\theta_0|)
	 \, \Rightarrow \, \frac{\lambda}{|\theta_0| + \epsilon} |\theta| \, ,
\end{equation}
therefore, with $\theta_0$ fixed, we consider only the last term above for optimization w.r.t.\ $\theta$.
Suppose we have a ``good'' initial solution $\{ \bar{\bm{\Phi}} \}$ to the problem (from e.g., using $L_{SGL} (\{ \bm{\Phi} \})$ instead of $L_{LSP} (\{ \bm{\Phi} \})$). Then, given $\{ \bar{\bm{\Phi}} \}$, a local linear approximation to ${P}(\{ \bm{\Phi} \})$ yields the convex function to be minimized
\begin{align}  
   &\tilde{L}_{LSP} (\{ \bm{\Phi} \})  =  G(\{ \bm{\Phi} \}, \{ \bm{\Phi}^\ast \}) + \tilde{P}(\{ \bm{\Phi} \}) , \label{eqth2_20f} \\
	 &  \tilde{P} (\{ \bm{\Phi} \} )   =  \sum_{k=1}^M \; \sum_{i \ne j}^p \lambda_{1ij} \,
		  \Big| [ {\bm{\Phi}}_k ]_{ij} \Big|  
		    +  \sum_{ i \ne j}^p \; {\lambda}_{2ij} \, \| {\bm{\Phi}}^{(ij)} \|  \label{eqth2_20g} \\
 &   \lambda_{1ij} = \bar{\lambda}_1 / (\big| [ \bar{\bm{\Phi}}_k ]_{ij} \big| + \epsilon) \, , \;\;
     \lambda_{2ij} = \bar{\lambda}_2 / (\| \bar{\bm{\Phi}}^{(ij)} \| + \epsilon)  \, .  \label{eqth2_20e}
\end{align}
This is then quite similar to adaptive lasso \cite{Zou2006} or adaptive sparse group lasso \cite{Tugnait21}, except that \cite{Zou2006} and \cite{Tugnait21} have $\epsilon = 0$. 
If we initialize with $[ \bar{\bm{\Phi}}_k ]_{ij} = 0$ for all $i \ne j$, we obtain an SGL cost.
\vspace*{-0.1in}
\section{Optimization}
Minimization of $L_{LSP} (\{ \bm{\Phi} \})$ is done iteratively where in each iteration, we minimize $\tilde{L}_{LSP}(\{ \bm{\Phi} \})$. Initially we take $\lambda_{1ij} = \bar{\lambda}_1$ and $\lambda_{2ij} = \bar{\lambda}_2$ $\forall \, (i,j)$. The result of this iteration is then used as $\{ \bar{\bm{\Phi}} \}$ in (\ref{eqth2_20e}) for next iteration.  
To optimize $\tilde{L}_{LSP}(\{ \bm{\Phi} \})$, using variable splitting, we first reformulate as in \cite{Tugnait18c}: 
\begin{align}  \label{eqth2_21}
 \min_{ \{\bm{\Phi} \}, \{ {\bm W} \} } & \Big\{ G(\{ \bm{\Phi} \}, \{ \bm{\Phi}^\ast \})
						+ P(\{ {\bm W} \})  \Big\} \;\;
\end{align}
subject to ${\bm W}_k = \bm{\Phi}_k \succ {\bf 0}, \; k=1,2, \cdots , M$, where $\{ \bm{\Phi} \} = \{\bm{\Phi}_k, \; k=1,2, \cdots , M\}$ and $\{ {\bm W} \} = \{{\bm W}_k, \; k=1,2, \cdots , M \}$.
In ADMM, we consider the scaled augmented Lagrangian for this problem \cite{Boyd2010, Danaher2014} 
\begin{align}
 L_\rho(\{ \bm{\Phi} \}, \{{\bm W} \}, \{{\bm U} \} ) = &  
   G(\{ \bm{\Phi} \}, \{ \bm{\Phi}^\ast \})
						+ \tilde{P}(\{ {\bm W} \})  \nonumber  \\
					&	+ \frac{\rho}{2} \sum_{k=1}^M \| \bm{\Phi}_k - {\bm W}_k + {\bm U}_k \|^2_F
\end{align}
where $\{ {\bm U} \} = \{{\bm U}_k, \; k=1,2, \cdots , M \}$ are dual variables, and $\rho >0$ is the ``penalty parameter'' \cite{Boyd2010}.

\subsection{ADMM Algorithm}
Given the results $\{ \bm{\Phi}^{(m)} \}, \{{\bm W}^{(m)} \}, \{{\bm U}^{(m)} \}$ of the $m$th iteration, in the $(m+1)$st iteration, an ADMM algorithm executes the following three updates:
\begin{itemize}[wide, labelwidth=!, labelindent=0pt]
\setlength{\itemindent}{0.04in}
\item[(a)] $\{ \bm{\Phi}^{(m+1)} \} \leftarrow \arg \min _{\{ \bm{\Phi} \}} 
            L_\rho ( \{ \bm{\Phi} \}, \{{\bm W}^{(m)} \}, \{{\bm U}^{(m)} \} )$
\item[(b)] $\{ {\bf W}^{(m+1)} \} \leftarrow \arg \min _{\{ {\bm W} \}} 
            L_\rho(\{ \bm{\Phi}^{(m+1)} \}, \{{\bm W} \}, \{{\bm U}^{(m)} \} )$
\item[(c)] $\{ {\bm U}^{(m+1)} \} \leftarrow \{ {\bm U}^{(m)} \}  +
   \left( \{ \bm{\Phi}^{(m+1)} \}  - \{ {\bm W}^{(m+1)} \} \right)$
\end{itemize}

\noindent{\bf Update (a)}: Notice that up to some terms not dependent upon $\bm{\Phi}_k$'s, $L_\rho ( \{ \bm{\Phi} \}, \{{\bm W}^{(m)} \}, \{{\bm U}^{(m)} \} ) =  \sum_{k=1}^M \frac{1}{2} L_{\rho k} (  \bm{\Phi}_k , {\bm W}_k^{(m)}, {\bm U}_k^{(m)} )$ (i.e., it is separable in $k$), where
\begin{align}  
	   L_{\rho k} & (  \bm{\Phi}_k , {\bm W}_k^{(m)}, {\bm U}_k^{(m)} )  :=  - \ln |\bm{\Phi}_k| - \ln |\bm{\Phi}_k^\ast| 
		        + {\rm tr} \Big( \hat{\bm S}_k \bm{\Phi}_k  \nonumber \\
					& \quad + 	     \hat{\bm S}_k^\ast \bm{\Phi}_k^\ast \Big)
	  + \rho \| \bm{\Phi}_k - {\bm W}_k^{(m)} + {\bm U}_k^{(m)} \|^2_F \, .
\end{align}
The solution to $\arg \min_{ \bm{\Phi}_k} 
            L_{\rho k} (  \bm{\Phi}_k , {\bm W}_k^{(m)}, {\bm U}_k^{(m)} )$ is as follows \cite[Sec.\ 6.5]{Boyd2010},\cite{Danaher2014, Tugnait18c}. Let ${\bm V}{\bm D}{\bm V}^H$ denote the eigen-decomposition of the matrix $(\hat{\bm S}_k-\rho {\bm W}_k^{(m)} + \rho {\bm U}_k^{(m)})$. Then $\bm{\Phi}_k^{(m+1)} = {\bm V} \tilde{\bm D} {\bm V}^H$ where $\tilde{\bm D}$ is the diagonal matrix with $\ell$th diagonal element
\begin{equation}  \label{eqth2_2004}
     \tilde{\bm D}_{\ell \ell} = \frac{1}{2 \rho} \left( -{\bm D}_{\ell \ell} + \sqrt{ |{\bm D}_{\ell \ell}|^2 + 4 \rho } \, \right) 
\end{equation}
By construction $\tilde{\bm D}_{\ell \ell} > 0$ for any $\rho > 0$, hence, $\bm{\Phi}_k^{(m+1)} = {\bm V} \tilde{\bm D} {\bm V}^H \succ 0$.

\noindent{\bf Update (b)}: Update $\{ {\bm W}_k^{(m+1)} \}_{k=1}^M$ as the minimizer of 
\begin{equation} \label{penaltyopt}
  \frac{\rho}{2} \sum_{k=1}^M \| {\bm W}_k - ( \bm{\Phi}_k^{(m+1)} + {\bm U}_k^{(m)} ) \|^2_F + \tilde{P}(\{ {\bm W} \}) 
\end{equation}
w.r.t.\ $\{ {\bm W} \}_{k=1}^M$. The solution follows from \cite{Tugnait18c} (which follows real-valued results of \cite{Friedman2010a}). Define $(b)_+ := \max(0,b)$, soft-thresholding operator $S(b,\beta) := (1-\beta/|b|)_+ b$, and vector operator $[{\bm S}({\bm a}, \beta)]_j = S(a_j,\alpha)$, $a_j = [{\bm a}]_j$. Let ${\bm A}_k = \bm{\Phi}_k^{(m+1)} + {\bm U}_k^{(m)}$. The solution to minimization of (\ref{penaltyopt}) is
\[
  [\hat{\bm W}_k]_{ij} = \left\{ \begin{array}{l}
	    [{\bm A}_k]_{ii} \, , \quad\quad \mbox{if } i=j \\
			S([{\bm A}_k]_{ij}, \frac{\lambda_{1ij}}{\rho}) \left( 1 - \frac{\lambda_{2ij}}
			{ \rho \| {\bm S}({\bm A}_k^{(ij)}, \lambda_{1ij}/\rho)\|}
			   \right)_+ \\
				\quad\quad\quad\quad \mbox{if } i \neq j  \end{array} \right.
\]

\noindent{\bf Update (c)}: For the scaled Lagrangian formulation of ADMM \cite{Boyd2010}, for $k=1,2, \cdots , M$, update ${\bm U}_k^{(m+1)}={\bm U}_k^{(m)} +( \bm{\Phi}_k^{(m+1)} - {\bm W}_k^{(m+1)})$.

\noindent{\bf Edge Selection}: Denote the converged estimates as $\hat{\bm{\Phi}}_k$, $k=1, \cdots , M$. If $\| \hat{\bm \Phi}^{(ij)} \| > 0$ then $\{ i,j \} \in {\cal E}$, else $\{ i,j \} \not\in {\cal E}$, $i \ne j$.

\vspace*{-0.1in}
\subsection{BIC for selection of $\lambda$ and $\alpha$}  
Given $n$, $K$ and $M$, the Bayesian information criterion (BIC) is given by
${\rm BIC}(\lambda , \alpha) =  2K  \sum_{k=1}^M \left( -\ln |\hat{\bm{\Phi}}_k|  + {\rm tr} \left( \hat{\bm S}_k  \hat{\bm{\Phi}}_k  \right) \right)   + \ln (2 K M) \, \sum_{k=1}^M (\mbox{\# of nonzero elements in } \hat{\bm{\Phi}}_k ) $ 
where $2KM$ are total number of real-valued measurements in frequency-domain and $2K$ are number of real-valued measurements per frequency point, with total $M$ frequencies in $(0, \pi)$. Pick $\alpha$ and $\lambda$ to minimize BIC. We use BIC to first select $\lambda$ over a grid of values with fixed $\alpha$, and then with selected $\lambda$, we search over $\alpha$ values in $[0,0.3]$. We search over $\lambda$ in the range $[\lambda_\ell , \lambda_u]$ selected via a heuristic as in \cite{Tugnait21}. For $\alpha = \alpha_0$ (=0.1), find the smallest $\lambda$, labeled $\lambda_{sm}$, for which we get a no-edge model; then we set $\lambda_u = \lambda_{sm}/2$ and $\lambda_\ell = \lambda_u/10$.  
\vspace*{-0.1in}

\section{Consistency} \label{consist}
 Define $p \times (pM)$ matrix ${\bm \Omega}$ as
\begin{equation}  \label{neqn100}
  {\bm \Omega} =[ {\bm \Phi}_1 \;  {\bm \Phi}_2 \; \cdots \; {\bm \Phi}_M ] \, ,
\end{equation}
and denote $L_{LSP}(\bm{\Omega}) = L_{LSP}(\{ \bm{\Phi} \})$, given by (\ref{eqth2_20e}). Set $\bar{\lambda}_1 = \alpha \lambda$ and $\bar{\lambda}_2 = (1-\alpha) \lambda$, with $0 \le \alpha \le 1$.
We now allow $p$, $M$, $K$ (see (\ref{window})), and $\lambda$ to be functions of sample size $n$, denoted as $p_n$, $M_n$, $K_n$ and $\lambda_n$, respectively. Note that $K_n M_n \approx n/2$. Pick $K_n = a_1 n^\gamma$ and $M_n = a_2 n^{1-\gamma}$ for some $0.5 < \gamma < 1$ so that $M_n/K_n \rightarrow 0$ as $n \rightarrow \infty$.  
Assume
\begin{itemize}
\setlength{\itemindent}{0.02in}
\item[(A1)] The $p-$dimensional time series $\{ {\bm x}(t) \}_{t=-\infty}^{\infty}$ is zero-mean stationary, Gaussian, satisfying $\sum_{\tau = -\infty}^\infty | [{\bm R}_{xx}( \tau )]_{k \ell} | < \infty$ for every $k, \ell \in V$.

\item[(A2)] Denote the true edge set of the graph by ${\cal E}_0$, implying that ${\cal E}_0 = \{ \{i,j\} ~:~ [{\bm S}^{-1}_{0}(f)]_{ij} \not\equiv 0, ~i\ne j,  ~ 0 \le f \le 0.5 \}$ where ${\bm S}_0(f) $ denotes the true PSD of ${\bm x}(t)$.
(We also use $\bm{\Phi}_{0k} $ for ${\bm S}^{-1}_{0}(\tilde{f}_k)$ where $\tilde{f}_k$ is as in (\ref{window}), and use ${\bm \Omega}_0$ to denote the true value of ${\bm \Omega}$). Assume that card$({\cal E}_0) =|({\cal E}_0)| \le s_{n0}$.

\item[(A3)] The minimum and maximum eigenvalues of $p_n \times p_n$ PSD ${\bm S}_0(f)  \succ {\bm 0}$  satisfy 
$0 < \beta_{\min}  \le \min_{f \in [0,0.5]} \phi_{\min}({\bm S}_0(f))$ and $\max_{f \in [0,0.5]} \phi_{\max}({\bm S}_0(f)) \le \beta_{\max} < \infty$. Here $\beta_{\min}$ and $\beta_{\max}$ are not functions of $n$ (or $p_n$).
\end{itemize}

Let $\hat{\bm{\Omega}}_\lambda = \arg\min_{\bm{\Omega} \,:\, \bm{\Phi}_{k} \succ {\bm 0}}  L_{LSP}(\bm{\Omega})$.
Theorem 1 whose proof is omitted for lack of space, establishes local consistency of $\hat{\bm{\Omega}}_\lambda$. \\
{\it Theorem 1 (Consistency)}. For $\tau > 2$, let
\begin{equation}  \label{naeq58}
   C_0 = 80 \, \max_{\ell, f} ( [{\bm S}_0(f)]_{\ell \ell}) 
	    \sqrt{ \frac{2 \ln (16 p_n^\tau M_n )}{ \ln ( p_n ) } } \, .
\end{equation}
Given any real numbers $\delta_1 \in (0,1)$, $\delta_2 > 0$ and $C_1 > 1$, let
\begin{align}  
    R = & C_{2} C_0 / \beta_{\min}^2 , \quad 
		   C_{2} = 2(2+C_{1}+ \delta_2)(1+\delta_1)^2  \, , \label{neq15ab0} \\
    r_n = & \sqrt{ \frac{M_n( p_n+ s_{n0}) \ln (p_n )}{K_n}}, \quad C_2 r_n = o(1)\, , \label{neq15ab1} \\
    N_1 = & \arg \min \left\{ n \, : \, K_n > 2 \ln (16 p_n^\tau M_n) \right\} \, ,  \label{neq15ab2} \\
		N_2 = & \arg \min \left\{ n \, : \, r_n \le 
		    \max \left( \frac{ \delta_1 }{\beta_{\min} R }, \frac{ \epsilon (C_1-1) }{ R } \right) \right\} \, . \label{neq15ab3}
\end{align}
Suppose the regularization parameter $\lambda_n$ and $\alpha \in [0,1]$ satisfy 
\begin{align}  
    C_0 C_1 & \sqrt{\frac{\ln (p_n )}{K_n}}  \le \frac{\lambda_n}{\epsilon \sqrt{M_n}}   \nonumber \\
		 & \le \frac{C_1  C_0 }{1+\alpha (\sqrt{M_n}-1)} 
		 \sqrt{ \left(1+ \frac{p_n}{s_{n0}} \right) \frac{\ln (p_n )}{K_n} } \, . \label{neq15abc}
\end{align}
Under assumptions (A1)-(A3), for sample size $n >  \max \{ N_1, N_2 \}$,
there exists a local minimizer $\hat{\bm{\Omega}}_\lambda$ such that
\begin{equation}  \label{neq15}
  \| \hat{\bm{\Omega}}_\lambda - \bm{\Omega}_0 \|_F 
	        \le R  r_n
\end{equation}
with probability greater than $1-1/p_n ^{\tau-2}$. A sufficient condition for the lower bound in (\ref{neq15abc}) to be less than the upper bound for every $\alpha \in [0,1]$ is  $C_1 = 2(1+\alpha (\sqrt{M_n}-1))$. In terms of rate of convergence,  
$\| \hat{\bm{\Omega}}_\lambda - \bm{\Omega}_0 \|_F = {\cal O}_P \left( C_1 r_n  \right) $ $\quad \bullet$
	
{\bf Remark 1}. Proof of Theorem 1 is patterned after \cite[Theorem 1]{Tugnait20} pertaining to the SGL approach; in turn, \cite{Tugnait20} exploits \cite{Rothman2008, Tugnait19c}. Convergence in \cite[Theorem 1]{Tugnait20} is global while here it is only local since LSP is non-convex. Assumption (A1) is needed to invoke \cite[Theorem 4.4.1]{Brillinger} to establish asymptotic statistics of DFT ${\bm d}_x(f_m)$'s; it is needed in \cite[Theorem 1]{Tugnait20} also where it was not explicitly stated. Assumptions (A2) and (A3) here are assumptions (A1) and (A2), respectively, in \cite{Tugnait20}. We use $C_1 >1$ to bound $\ln(1+x) \ge x/C_1$ for $0 \le x \le C_1-1$. $\;\; \Box$

\section{Numerical Examples} All ADMM approaches used variable penalty parameter $\rho$ as in \cite[Sec.\ 3.4.1]{Boyd2010}, and the stopping (convergence) criterion following \cite[Sec.\ 3.3.1]{Boyd2010}. We picked $\epsilon = 0.0001$ in LSP.

{\bf Synthetic Data}: Consider $p=128$, 16 clusters (communities) of 8 nodes each, where nodes within a community are not connected to any nodes in other communities. Within any community of 8 nodes, the data are generated using a vector autoregressive (VAR) model of order 3. Consider community $q$, $q=1,2, \cdots , 16$. Then  ${\bm x}^{(q)}(t) \in \mathbb{R}^8$ is generated as
\[
    {\bm x}^{(q)}(t) = \sum_{i=1}^3 {\bm A}^{(q)}_i {\bm x}^{(q)}(t-i) + {\bm w}^{(q)}(t) 
\] 
with  ${\bm w}^{(q)}(t)$ as i.i.d.\ zero-mean Gaussian with identity covariance matrix. Only 10\% of entries of ${\bm A}^{(q)}_i$'s are nonzero and the nonzero elements are independently and uniformly distributed over $[-0.8,0.8]$. We then check if the VAR(3) model is stable with all eigenvalues of the companion matrix $\le 0.95$ in magnitude; if not, we re-draw randomly till this condition is fulfilled. The overall data ${\bm x}(t)$ is given by ${\bm x}(t) = [\, {\bm x}^{(1) \top}(t) \; \cdots \; {\bm x}^{(16) \top}(t) \, ]^\top \in \mathbb{R}^{p}$. First 100 samples are discarded to eliminate transients. This set-up leads to approximately 3.5\% connected edges.

\begin{figure}[ht]
  \centering
  \includegraphics[width=0.8\linewidth]{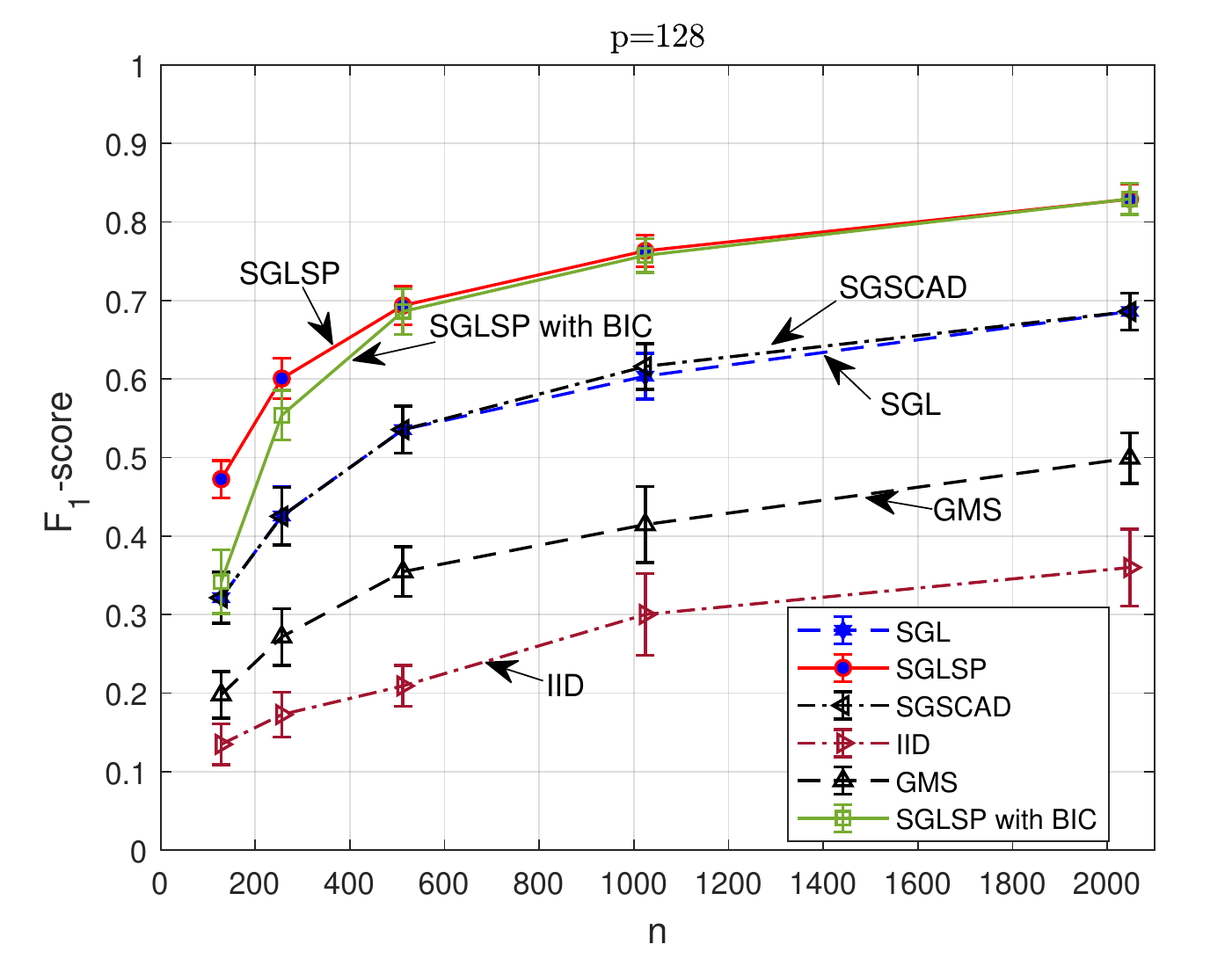} 
	\vspace*{-0.1in}
  \caption{$F_1$-scores based on 100 runs for 6 approaches. SGLSP: proposed sparse-group log-sum penalty approach; SGL: sparse-group lasso \cite{Tugnait18c, Tugnait20}; SGLSCAD: sparse-group SCAD-penalized method optimized using ADMM, where LSP is replaced with the non-convex SCAD penalty \cite{Lam2009}; IID: exploits only ${\bm R}_{xx}(0)$; GMS: method of \cite{Jung2015a}; SGLSP with BIC: proposed SGLSP with tuning parameters selected via BIC.}
  \label{fig1}
\end{figure}

Simulation results based on 100 runs are shown in Fig.\ \ref{fig1}. We used $M=4$ for all samples sizes $n=128,256,512,1024, 2048$ with corresponding $K=15,31,63,127, 255$, respectively.  The performance measure is $F_1$-score for efficacy in edge detection. The $F_1$-score is defined as $F_1 = 2 \times \mbox{precision} \times \mbox{recall}/(\mbox{precision} + \mbox{recall})$ where $\mbox{precision} = | \hat{\cal E} \cap {\cal E}_0|/ |\hat{\cal E}|$, $\mbox{recall} = |\hat{\cal E} \cap {\cal E}_0|/ |{\cal E}_0|$, and ${\cal E}_0$ and $ \hat{\cal E}$ denote the true and estimated edge sets, respectively. 
Six approaches were tested: {\bf (i)} Proposed sparse-group log-sum penalty based approach, labeled ``SGLSP'' in Fig.\ \ref{fig1}. {\bf (ii)} Sparse-group lasso (labeled ``SGL'') of \cite{Tugnait18c, Tugnait20} optimized using ADMM as in this paper (this approach also initializes SGLSP). {\bf (iii)} Sparse-group SCAD-penalized method (labeled ``SGSCAD'') optimized using ADMM, where instead of LSP we use the non-convex SCAD penalty (see \cite{Lam2009} for SCAD). {\bf (iv)} An i.i.d.\ modeling approach that exploits only the sample covariance $\frac{1}{n} \sum_{t=0}^{n-1} {\bm x}(t) {\bm x}^\top(t)$ (labeled ``IID''), implemented via the ADMM (adaptive) lasso approach (\cite[Sec.\ 6.4]{Boyd2010}). In this approach edge $\{i,j\}$ exists in the CIG iff $\Omega_{ij} \ne 0$ where precision matrix ${\bm \Omega} = {\bm R}_{xx}^{-1}(0)$. {\bf (v)} The ADMM approach of \cite{Jung2015a}, labeled ``GMS'' (graphical model selection), which was applied with $F=4$ (four frequency points, corresponds to our $M=4$) and all other default settings of \cite{Jung2015a} to compute the PSDs. The tuning parameters, $(\alpha, \lambda)$ for SGLSP, SGL and SGSCAD, and lasso parameter $\lambda$ for IID and GMS, were selected via an exhaustive search over a grid of values to maximize the $F_1$-score (which requires knowledge of the true edge-set). The results shown in Fig.\ \ref{fig1} are based on these optimized tuning parameters. {\bf (vi)} The sixth approach labeled ``SGLSP with BIC'' in Fig.\ \ref{fig1}, denotes our proposed ADMM approach for SGLSP where $(\alpha, \lambda)$ were selected via BIC, not requiring knowledge of the ground-truth.

It is seen that with $F_1$-score as the performance metric, our proposed SGLSP method significantly outperforms other approaches. The BIC-based SGLSP method yields performance that is close to that based on optimized parameter selection. SGLSCAD offers very little improvement over SGL, in contrast with proposed SGLSP.

\begin{figure}[ht]
\begin{subfigure}[t]{.24\textwidth}
  \centering
  \includegraphics[width=\linewidth]{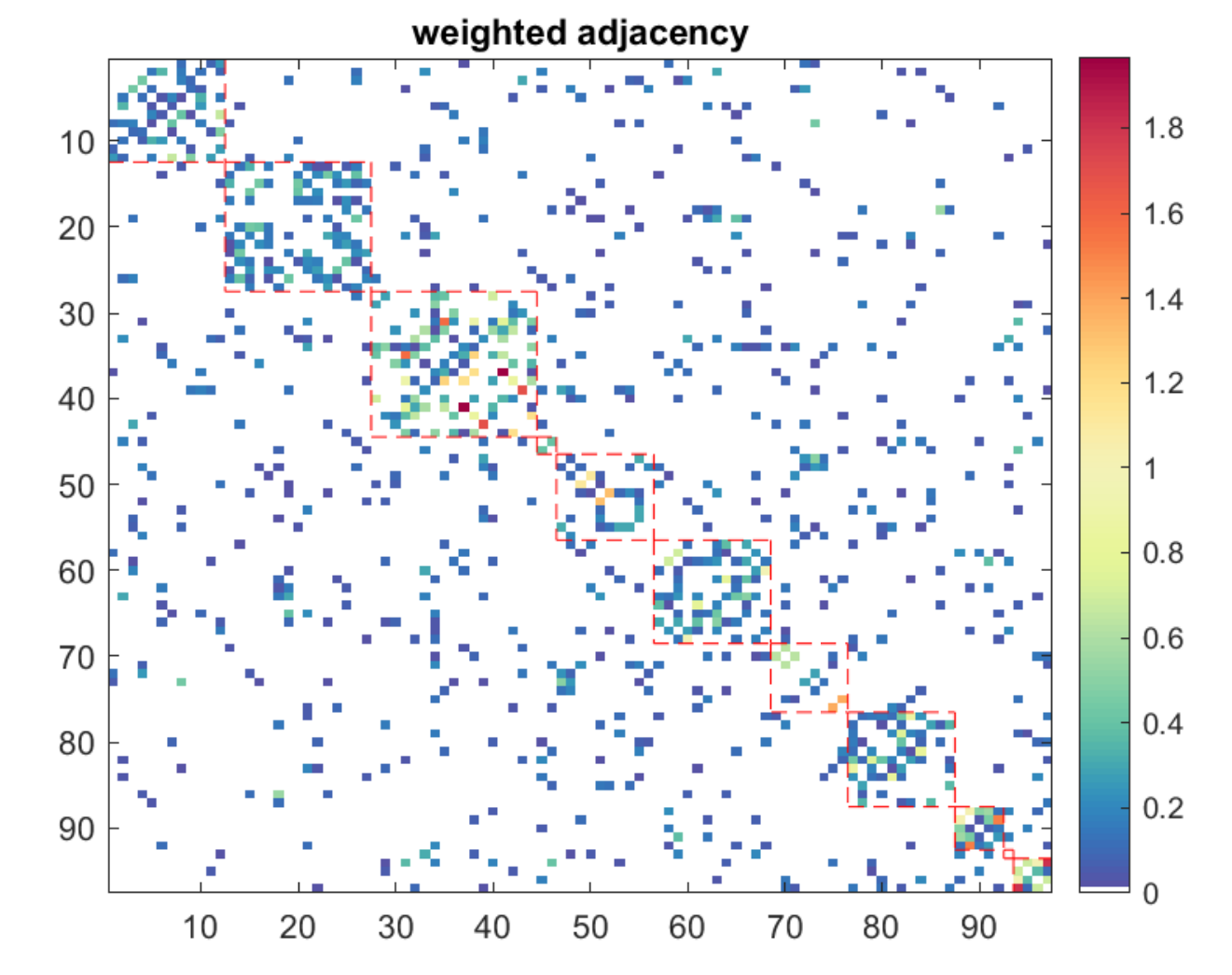}  
  \caption{Estimated $|\Omega_{ij}|$ as edge weight; \\ \hspace*{0.05in}  579 edges.}
  \label{fig3a}
\end{subfigure}%
\begin{subfigure}[t]{.24\textwidth}
  \centering
  \includegraphics[width=\linewidth]{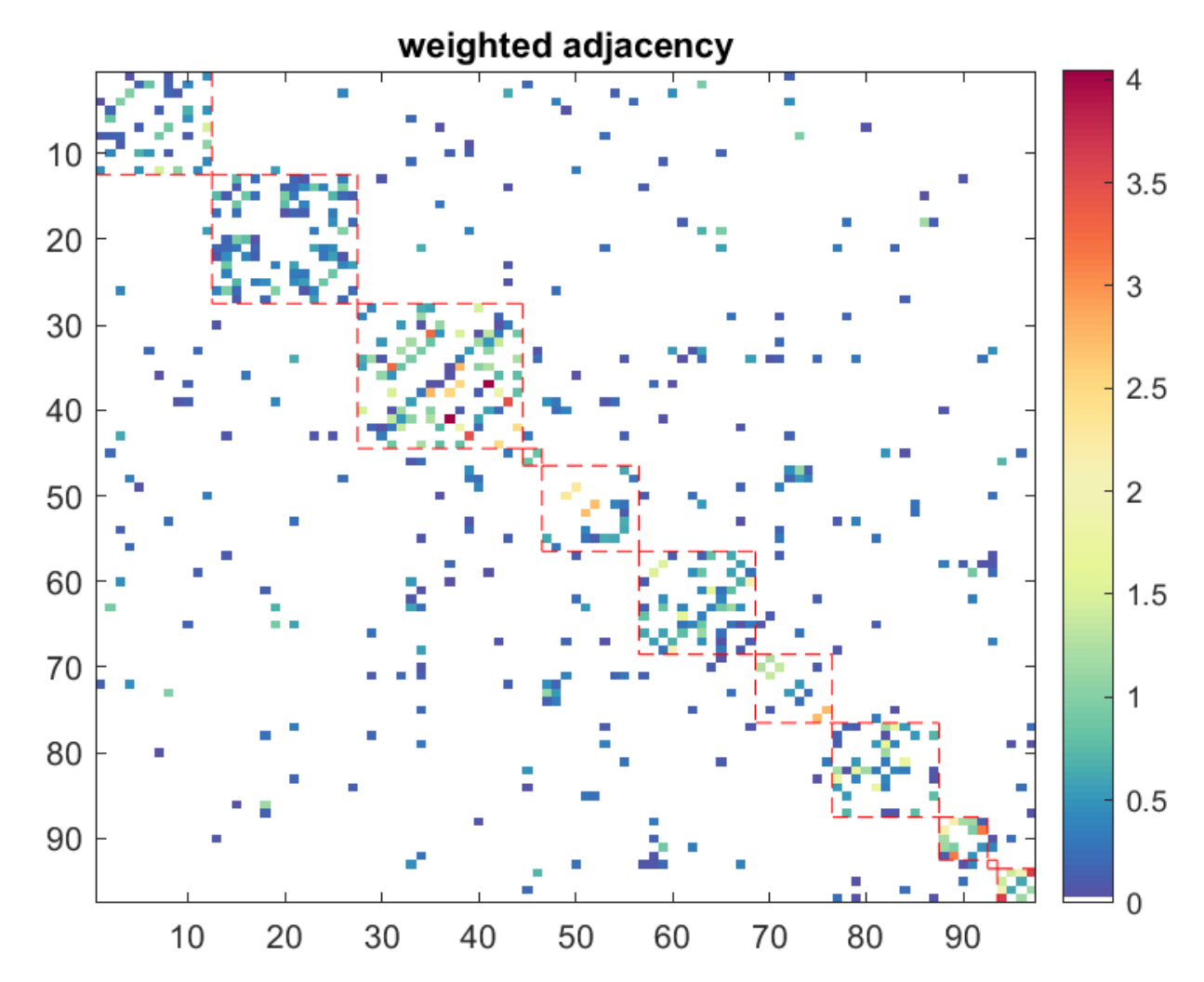}  
  \caption{Estimated $\sqrt{\sum_{k=1}^M | [\hat{\bm{\Phi}}_k]_{ij} |^2 }$ \\ \hspace*{0.05in}  as edge weight; 321 edges.}
  \label{fig3b}
\end{subfigure}
\vspace*{-0.1in}
\caption{Weighted adjacency matrices for financial time series. The red squares (in dashed lines) show the 11 sectors -- they are not part of the adjacency matrices.} 
\label{fig3}
\end{figure}

{\bf Real data: Financial Time Series}: We consider daily share prices (at close of the day) of 97 stocks in S\&P 100 index from Jan. 1, 2013 through Jan.\ 1, 2018, yielding 1259 samples. If $y_m(t)$ is share price of $m$th stock on day $t$, we consider (as is conventional in such studies) $x_m(t) = \ln (y_m(t)/y_m(t-1))$ as the time series to analyze, yielding $n=1258$ and $p=97$. These 97 stocks are classified into 11 sectors and we order the nodes to group them as information technology (nodes 1-12), health care (13-27), financials (28-44), real estate (45-46), consumer discretionary (47-56), industrials (57-68), communication services (69-76), consumer staples (77-87), energy (88-92), materials (93), utilities (94-97). The weighted adjacency matrices resulting from the IID modeling approach (estimated $|\Omega_{ij}|$ is the edge weight), and the proposed SGLSP (with BIC) approach with $M=4$, ($\sqrt{\sum_{k=1}^M | [\hat{\bm{\Phi}}_k]_{ij} |^2 }$ is the edge weight), are shown in Fig.\ \ref{fig3}. In both cases we used BIC to determine the tuning parameters. While the ground truth is unknown, the dependent time series based proposed approach yields sparser, more interpretable CIG (321 edges for the proposed approach versus 579 edges for IID modeling) which also conforms better with the sector classification according to the Global Industry Classification Standard.
\vspace*{-0.1in}

\vspace*{-0.3in}
\vspace*{-0.2in}

\section{Conclusions} Sparse-group lasso penalized log-likelihood approach in frequency-domain has been considered in \cite{Tugnait18c, Tugnait20} for graph learning for dependent time series. In this paper we considered a sparse-group log-sum penalty instead of the sparse-group lasso (SGL) penalty to regularize the problem. An ADMM approach for iterative optimization of the non-convex problem was presented. We provided sufficient conditions for consistency of a local estimator of inverse PSD. We illustrated our approach using numerical examples utilizing both synthetic and real data.
Synthetic data example showed that our SGLSP approach significantly outperformed the SGL and other approaches in correctly detecting the graph edges with $F_1$-score as performance metric.

\bibliographystyle{unsrt} 

\end{document}